\documentclass[acmtog,nonacm]{acmart}

\setcopyright{none}
\renewcommand\footnotetextcopyrightpermission[1]{}
\usepackage{booktabs} 
\usepackage{enumitem}

\usepackage[ruled]{algorithm2e} 

\SetAlFnt{\small}
\SetAlCapFnt{\small}
\SetAlCapNameFnt{\small}
\SetAlCapHSkip{0pt}

\begin{document}
\title{FW-VTON: Flattening-and-Warping for Person-to-Person Virtual Try-on}

\author{Zheng Wang}
\authornote{Both authors contributed equally to this research.}
\affiliation{%
  \institution{Shanghai Jiao Tong University}
  \city{Shanghai}
  \country{China}}
\email{wang.zheng@sjtu.edu.cn}

\author{Xianbing Sun}
\authornotemark[1]
\affiliation{%
  \institution{Shanghai Jiao Tong University}
  \city{Shanghai}
  \country{China}}
\email{fufengsjtu@sjtu.edu.cn}

\author{Shengyi Wu}
\affiliation{%
  \institution{Shanghai Jiao Tong University}
  \city{Shanghai}
  \country{China}}
\email{wsykk2@sjtu.edu.cn}

\author{Jiahui Zhan}
\affiliation{%
  \institution{Shanghai Jiao Tong University}
  \city{Shanghai}
  \country{China}}
\email{jiahuizhan@sjtu.edu.cn}

\author{Jianlou Si}
\affiliation{%
  \institution{TeleAI, China Telecom}
  \country{China}}
\email{sijianlou@gmail.com}

\author{Chi Zhang}
\affiliation{%
  \institution{TeleAI, China Telecom}
  \country{China}}
\email{zhangc120@chinatelecom.cn}

\author{Liqing Zhang}
\affiliation{%
  \institution{Shanghai Jiao Tong University}
  \city{Shanghai}
  \country{China}}
\email{zhang-lq@cs.sjtu.edu.cn}

\author{Jianfu Zhang}
\authornote{Corresponding author.}
\affiliation{%
  \institution{Shanghai Jiao Tong University}
  \city{Shanghai}
  \country{China}}
\email{c.sis@sjtu.edu.cn}

\renewcommand\shortauthors{Zheng Wang,Xianbing Sun. et al}
\begin{teaserfigure}
  \includegraphics[width=\textwidth]{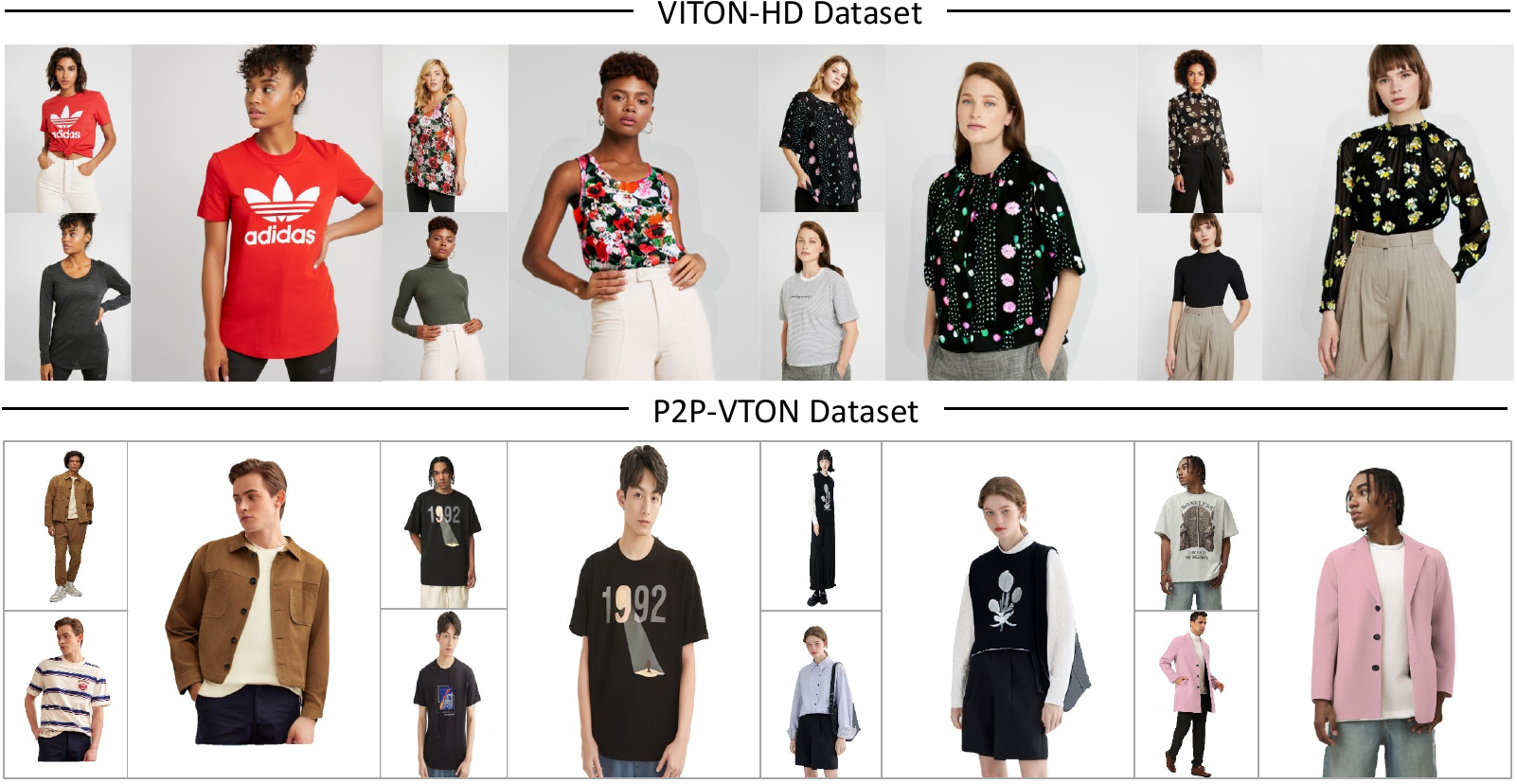}
  \caption{Examples of results from FW-VTON for person-to-person virtual try-on task.}
  \label{fig:first}
\end{teaserfigure}

\begin{abstract}
    Traditional virtual try-on methods primarily focus on the garment-to-person try-on task, which requires flat garment representations. In contrast, this paper introduces a novel approach to the person-to-person try-on task. Unlike the garment-to-person try-on task, the person-to-person task only involves two input images: one depicting the target person and the other showing the garment worn by a different individual. The goal is to generate a realistic combination of the target person with the desired garment. To this end, we propose Flattening-and-Warping Virtual Try-On (\textbf{FW-VTON}), a method that operates in three stages: (1) extracting the flattened garment image from the source image; (2) warping the garment to align with the target pose; and (3) integrating the warped garment seamlessly onto the target person. To overcome the challenges posed by the lack of high-quality datasets for this task, we introduce a new dataset specifically designed for person-to-person try-on scenarios. Experimental evaluations demonstrate that FW-VTON achieves state-of-the-art performance, with superior results in both qualitative and quantitative assessments, and also excels in garment extraction subtasks.
\end{abstract}

\maketitle

\section{Introduction}
Virtual Try-On (VTON) has emerged as a crucial technology in e-commerce \cite{bai2022single,choi2021viton,ge2021parser,gou2023taming,lee2022high,morelli2022dress,morelli2023ladi,choi2025improving,xu2024ootdiffusion}, enabling personalized garment visualization and significantly reducing the uncertainties of online shopping. Beyond e-commerce, VTON holds the potential to revolutionize real-world applications by providing tailored clothing visualization and enhancing user experiences across various domains.
The primary challenge in VTON lies in accurately transferring garments onto target images while preserving fine-grained details, such as textures, folds, and patterns. High-frequency features from the source image must be retained, and spatial and geometric consistency between the garment, target pose, and body shape is essential for generating realistic outputs.

To address these challenges, researchers have explored two major approaches: Generative Adversarial Networks \cite{goodfellow2020generative} and Diffusion Models \cite{ho2020denoising,rombach2022high}. GAN-based approaches \cite{choi2021viton,ge2021disentangled,lee2022high,xie2023gp} typically rely on warping the garment from the source image to align with the target person’s pose, followed by a fitting process to ensure proper alignment. However, GANs often struggle to produce high-quality, visually consistent results due to their inherent limitations in capturing fine-grained details.
Recent advancements in diffusion-based approaches \cite{gou2023taming,kim2024stableviton,zhu2023tryondiffusion,xu2024ootdiffusion,choi2025improving} have significantly improved the performance of VTON tasks, enabling the generation of more realistic images. These methods address some of the limitations of GANs by leveraging iterative denoising processes for fine detail preservation. Despite these improvements, challenges persist in effectively integrating garment details into the target image. Various methods tackle this by employing warping modules or reference net to assist in the integration process \cite{gou2023taming,kim2024stableviton,zhu2023tryondiffusion,xu2024ootdiffusion,choi2025improving,chong2024catvton}.

While most prior VTON methods focus on the garment-to-person task, we address a more complex and universally applicable challenge: the person-to-person VTON task. In the garment-to-person setting, the objective is to transfer a flat garment image onto a target person, leveraging both the garment and person images. In contrast, the person-to-person setting involves an additional layer of complexity, as the source image depicts a garment being worn by another person. This introduces occlusions and distortions in the garment image, making the task not only about transferring the garment to the target person but also about repairing and reconstructing the occluded and distorted portions of the garment.
To overcome these challenges, we propose Flattening-and-Warping Virtual Try-On (FW-VTON), which operates in three stages:
\textbf{(i)} Flattening Stage: The model takes the source person image as input and generates a flattened garment image, reconstructing the occluded regions; \textbf{(ii)} Warping Stage: The model warps the garment image to align with the pose of the target person; \textbf{(iii)} Integration Stage: Finally, the flattened and warped garment images are integrated to generate a realistic image of the target person wearing the garment.
A significant limitation in existing research is the lack of high-quality datasets for person-to-person scenarios, which hinders training and evaluation. To address this gap, we introduce a novel dataset specifically designed for the person-to-person setting, providing a robust foundation for the development and benchmarking of advanced VTON models.
Our proposed FW-VTON demonstrates superior performance in the person-to-person VTON task and achieves state-of-the-art results on the newly introduced P2P-VTON dataset.

Our contributions are summarized as follows:
\begin{itemize}[leftmargin=*]
\item We present the P2P-VTON dataset, specifically designed to facilitate the training and evaluation of VTON models in the person-to-person setting.
\item We introduce a novel method for extracting garment images from person images, significantly outperforming current state-of-the-art approaches focused solely on this subtask.
\item We propose a warping module that directly generates warped garment images by aligning the garment from the source image with the target person image, eliminating the dependency on flat garment images.
\item We demonstrate the effectiveness of our method through comprehensive experiments conducted on both the VITON-HD dataset \cite{choi2021viton} and our newly introduced P2P-VTON dataset, achieving state-of-the-art performance on the person-to-person VTON task.  
\end{itemize}

\begin{figure*}[t]
  \includegraphics[width=0.9\textwidth]{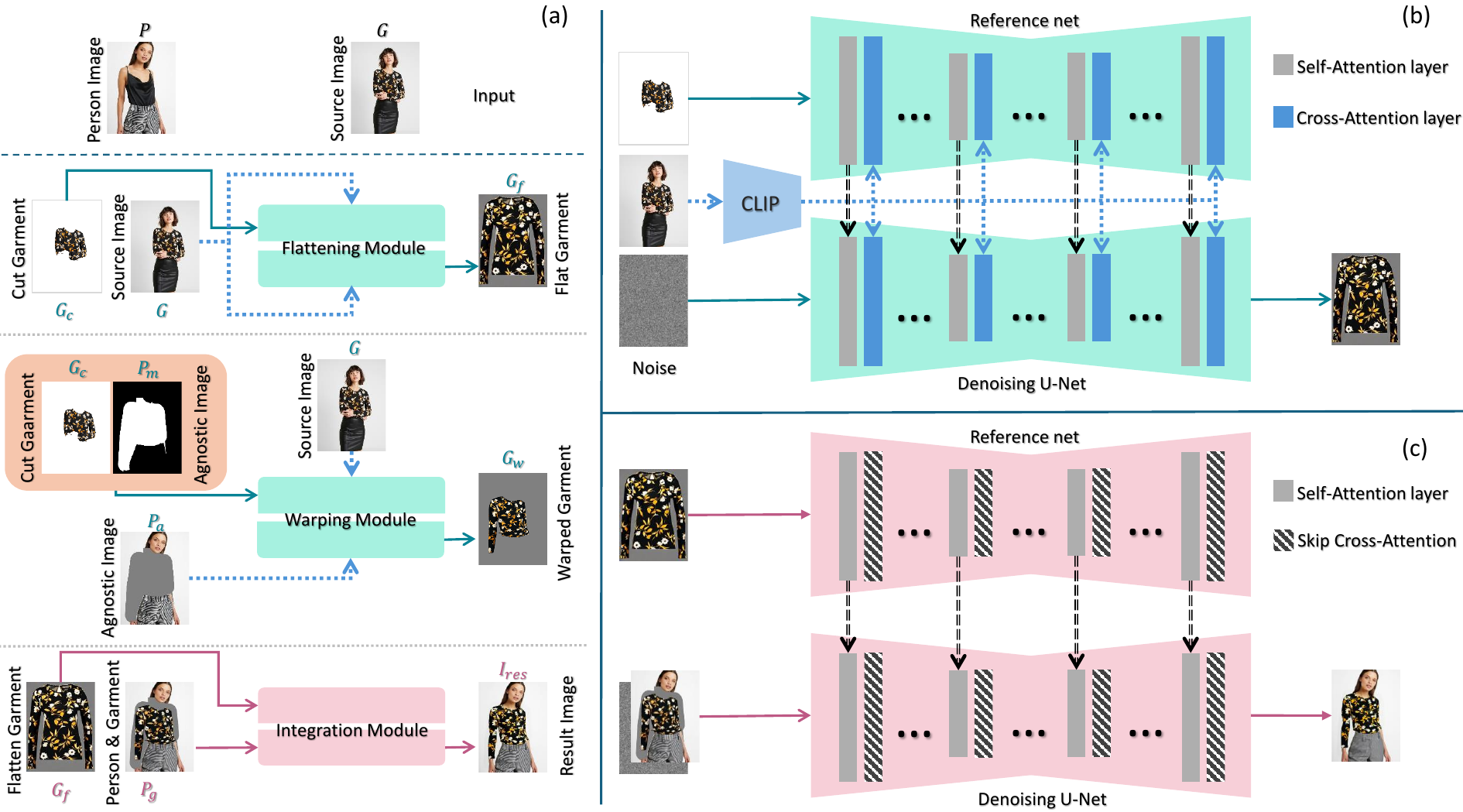}
  \caption{Overview of the proposed FW-VTON framework. The overall pipeline includes three stages as illustrated in (a).  (b) shows the detailed architecture of the Flattening Module based on a Dual U-Net; the Warping Module adopts the same architecture but with different inputs, as shown in the corresponding part of (a). In the final Integration Module illustrated in (c), the cross-attention layers are removed. The image $P_g$ provides structure guidance, while $G_f$ is processed through a reference net to supply fine details.  
  Solid arrows represent inputs encoded by a VAE before being passed to the U-Net, while dashed arrows denote structure features extracted by CLIP for cross-attention guidance(excluding the black dashed arrows).}
  \label{fig:overall}
\end{figure*}
\section{Related Works}

\subsection{Image-based VTON}
The task of image-based virtual try-on (VTON) involves generating a composite image of a target person wearing a desired garment, given a reference image of the person and a conditional image of the garment (or another person wearing it). Early methods~\cite{choi2021viton,ge2021disentangled,lee2022high,xie2023gp} primarily relied on \textit{Generative Adversarial Networks (GANs)}~\cite{goodfellow2020generative}. These approaches typically incorporated a warping module to deform the garment to match the target pose, followed by an image synthesis stage. While warping significantly improved garment alignment and realism, GAN-based methods often struggled with fine-detail preservation and generalization to unseen data.

More recently, diffusion models~\cite{ho2020denoising,rombach2022high} have emerged as a powerful alternative for VTON, offering superior generative quality. One of the core challenges in VTON is the accurate transfer of fine garment textures and structures onto the target body. DCI-VTON~\cite{gou2023taming} builds on GAN frameworks but enhances detail retention via improved warping. TryOnDiffusion~\cite{zhu2023tryondiffusion} adopts a dual U-Net architecture to better model garment features, though it relies heavily on large-scale training data due to the absence of strong priors. Stable-VTON~\cite{kim2024stableviton} utilizes the Stable Diffusion encoder~\cite{rombach2022high} as a garment encoder and combines it with a denoising U-Net for pose-aware synthesis.
Other notable works, such as IDM-VTON~\cite{choi2025improving} and OOTDiffusion~\cite{xu2024ootdiffusion}, introduce a second denoising U-Net—termed the reference net—to process garment-specific information, enhancing both texture preservation and integration quality. In contrast, CatVTON~\cite{chong2024catvton} questions the necessity of such additional components. It directly concatenates the garment and person images into the input channels of a Stable Diffusion model, omitting cross-attention mechanisms, and achieves competitive results with lower computational cost.

\subsection{Person-to-Person VTON}
The person-to-person VTON task is both more challenging and broadly applicable than the garment-to-person VTON task. A key distinction in this task is the inclusion of an additional ``try-off'' subtask: extracting garment information from a person-wearing image to generate a flattened garment image. This process, referred to as the ``flattening stage'', has recently gained attention.FLDM-VTON~\cite{wang2024fldm} is the first to propose this flatten-and-warp paradigm by introducing a clothes flattening network for training supervision, as a solution to the lack of pixel-level guidance in latent diffusion-based try-on models.TryOffDiff~\cite{velioglu2024tryoffdiff} later addresses this subtask by guiding Stable Diffusion to generate the flattened garment image through an image adaptor, which provides conditional information via the diffusion cross-attention layer. Similarly, TryOffAnyone~\cite{xarchakos2024tryoffanyone} employs an approach akin to CatVTON, concatenating all condition information in the input channel of Stable Diffusion and bypassing the cross-attention mechanism.

Alternatively, this subtask can be integrated directly into the VTON process, as demonstrated in prior works like TryOnDiffusion~\cite{zhu2023tryondiffusion} and CatVTON~\cite{chong2024catvton}, which combine garment extraction and VTON into a unified pipeline. Additionally, methods like IDM-VTON and OOTDiffusion adapt the input garment image by extracting the garment region from the person-wearing image, making them compatible with the person-to-person VTON task.

A significant limitation of current major VTON datasets, such as VITON-HD~\cite{choi2021viton} and DressCode~\cite{morelli2022dress}, is their exclusive focus on flat garment images paired with person-wearing images. While DeepFashion~\cite{liu2016deepfashion} includes images of people wearing the same garment, these images are taken from varying perspectives rather than poses, making them unsuitable for training person-to-person VTON models. To address this limitation, we introduce the P2P-VTON dataset, which provides a more suitable foundation for training VTON models that require diverse input data.

\section{Methodologies}
\label{sec:sim}
The person-to-person VTON task involves two inputs: a target person image, denoted as  $\mathcal{P}$, and a source image of a garment worn by another person, denoted as   $\mathcal{G}$. 
This setting introduces unique challenges compared to traditional garment-to-person VTON tasks, primarily due to the partial and deformed visibility of the garment in the source image. To address this added complexity, we decompose our approach into distinct stages (see Figure~\ref{fig:overall}).
As preprocessing steps, we extract essential semantic information through human parsing maps $\mathcal{S}_p$ and $\mathcal{S}_g$ corresponding to the target person $\mathcal{P}$ and the garment-wearing person $\mathcal{G}$, respectively. 
Additionally, we generate an agnostic image $\mathcal{P}_a$, which retains the target person's identity while removing the original garment.  
The parsing maps and the agnostic image are obtained using existing methods~\cite{li2020self,papandreou2017towards}. 
Our framework then processes these inputs through three sequential stages: flattening, warping, and integration. 
\paragraph{Flattening Stage.} This stage performs a critical role in reconstructing complete garment information from partial observations. Using the parsing map $\mathcal{S}_g$ of the garment-wearing person image $\mathcal{G}$, we isolate the garment worn by the person, obtaining the cut-out garment image $\mathcal{G}_c$. Then, we apply a flattening module to transform $\mathcal{G}_c$ into a fully reconstructed flattened garment image, denoted as $\mathcal{G}_f$, which retains all the original garment details.

\paragraph{Warping Stage.} This stage generates the warped garment to match the target person's physique and pose. This process integrates multiple inputs: the target agnostic image  $\mathcal{P}_a$ and garment region mask $\mathcal{P}_m$, the Source Image $\mathcal{G}$, and the cut-out garment image $\mathcal{G}_c$ to generate a warped garment $\mathcal{G}_w$ that aligns precisely with the target person's body structure and positioning. 

\paragraph{Integration Stage.} Finally, we merge the warped garment $\mathcal{G}_w$ with the target person image $\mathcal{P}$, resulting in the final output image $\mathcal{I}_{res}$, where the target person is shown wearing the warped garment $\mathcal{G}_w$.

\subsection{Flattening Module}
Given an image $\mathcal{G}$, the goal of the flattening module is to extract garment information and generate a corresponding flattened image $\mathcal{G}_f$, effectively reconstructing the complete garment despite challenges such as occlusions and pose variations.
Recent methods have addressed this task, including TryOffDiff~\cite{velioglu2024tryoffdiff}, which employs an adaptor to provide conditioning for diffusion models, and TryOffAnyone~\cite{xarchakos2024tryoffanyone}, which simplifies the pipeline by directly concatenating condition information into input channels—similar to CatVTON~\cite{chong2024catvton}—thus eliminating cross-attention layers.
Our approach leverages a reference network architecture~\cite{hu2024animate}, shown to be effective in related works~\cite{choi2025improving,xu2024ootdiffusion}, to guide the denoising U-Net in reconstructing garment details. To further address occluded or incomplete garment regions, we adopt a dual U-Net design, which balances reconstruction quality and computational efficiency. Extensive experiments demonstrate that this combination is particularly effective for the garment flattening subtask.
\paragraph{Dual U-Net.}
Our dual U-Net architecture consists of a reference net and a denoising U-Net, both based on the Stable Diffusion U-Net~\cite{rombach2022high} structure and initialized with its pre-trained weights. 
The reference net and denoising U-Net are integrated through the self-attention mechanism. 
Given the feature map ${x_d} \in \mathbb{R}^{H \times W \times C}$ from the denoising U-Net and ${x_r} \in \mathbb{R}^{H \times W \times C}$ from the reference net, we concatenate these feature maps along the spatial dimension (the $W$ dimension), resulting in a combined feature map ${x_c} \in \mathbb{R}^{H \times 2W \times C}$. 
After applying self-attention to this concatenated feature map, we selectively extract the first half of the processed features. These refined features are then propagated to subsequent layers of the denoising U-Net, enabling the network to leverage the complementary information from both pathways effectively. 

\paragraph{Using $\mathcal{G}$ for Cross-Attention.}
To effectively reconstruct the flattened garment image $\mathcal{G}_f$, our dual U-Net architecture implements a fine-grained encoding and attention mechanism. The process begins with the extraction of the cut garment $\mathcal{G}_c\in\mathbb{R}^{H \times W\times3}$ from the source image $\mathcal{G}$ using the human parsing map $\mathcal{S}_g$. 
This extracted garment is then encoded by Variational Autoencoder (VAE)~\cite{kingma2013auto}, resulting in the encoded representation $\mathcal{E}(\mathcal{G}_c)\in\mathbb{R}^{H\times W\times4}$, which serves as input to the reference net. 
While a straightforward approach might suggest using $\mathcal{G}_c$ directly with the CLIP~\cite{radford2021learning} encoder for cross-attention in both networks, our ablation studies demonstrate the limitations of this method.   
Analysis reveals that this suboptimal performance stems from insufficient pose context. We discovered that providing comprehensive body information enhances the network's ability to reconstruct garment details accurately. 
To address this limitation, we leverage the complete source image  $\mathcal{G}$ as input to the CLIP encoder rather than just the isolated garment. This modification provides essential contextual information about garment fit and flattening, resulting in significantly improved reconstruction quality in the final output. 

\subsection{Warping Module}
\label{subsection:warping}
The Warping Module addresses the distinct challenge of pose alignment, requiring a different approach from the flattening task. 
We maintain  $\mathcal{G}_c$ as conditional information and utilize the reference net for feature extraction, which guides the denoising U-Net in generating the final aligned garment. 

Our implementation begins with preprocessing the target person image  $\mathcal{P}$ to create an agnostic representation  $\mathcal{P}_a$. 
Using the human parsing map $\mathcal{S}_p$, we generate a garment region mask  $\mathcal{P}_m \in \mathbb{R}^{H \times W}$. 
We then resize $\mathcal{P}_m$ to $\mathcal{E}(\mathcal{P}_m) \in \mathbb{R}^{\frac{H}{8} \times \frac{W}{8}}$, and finally this resized mask is concatenated with the cut garment image $\mathcal{G}_c$, yielding ${x}_t \in \mathbb{R}^{H \times W \times 5}$. 
To accommodate this expanded input dimension, we modify the reference net's initial convolution layer by extending its input channels from 4 to 5 through zero-parameter padding. 
The final step involves providing $\mathcal{G}$ to the CLIP encoder for the reference net, while $\mathcal{P}_a$ serves as input to the CLIP encoder for the denoising U-Net, completing our dual U-Net warping architecture.

\subsection{Integration Module}
At the integration stage, we have already obtained both the warped garment image $\mathcal{G}_w$ and the flattened garment image $\mathcal{G}_f$. The remaining challenge lies in effectively combining them. In the virtual try-on task, the two core challenges are: (1) accurately aligning the garment with the body shape, and (2) faithfully preserving garment details.

To address these challenges, our integration module still adopts a dual U-Net architecture. The reference net takes the flattened garment $\mathcal{G}_f$ as input to provide high-fidelity garment detail information. Meanwhile, the denoising U-Net is conditioned on the warped garment $\mathcal{G}_w$, which encodes structural cues such as garment shape and the spatial layout of internal patterns (\textit{e.g.}, stripes, textures) as it would appear when worn.

To inject structural information into the denoising U-Net, we construct a composite image $\mathcal{P}_g \in \mathbb{R}^{H \times W \times 3}$ by overlaying $\mathcal{G}_w$ onto the agnostic person image $\mathcal{P}_a$. This composite is then passed through a VAE encoder to produce a latent representation $\mathcal{E}(\mathcal{P}_g) \in \mathbb{R}^{H \times W \times 4}$. We concatenate this latent with Gaussian noise $\epsilon \in \mathbb{R}^{H \times W \times 4}$, resulting in an 8-channel input to the denoising U-Net. To accommodate this, we modify the first convolutional layer of the diffusion model by introducing zero-padded parameters.

We also remove the cross-attention layers from both the reference net and the denoising U-Net. In Stable Diffusion~\cite{rombach2022high}, these layers are designed for image-text alignment. However, the try-on task uses only image-based conditioning, making cross-attention largely unnecessary. At best, it may offer weak structural cues, which are already provided explicitly via $\mathcal{G}_w$. Removing cross-attention reduces computational overhead and improves performance, as shown in the \textit{Supplementary Materials}.

\begin{figure}[t]
  \includegraphics[width=1\columnwidth]{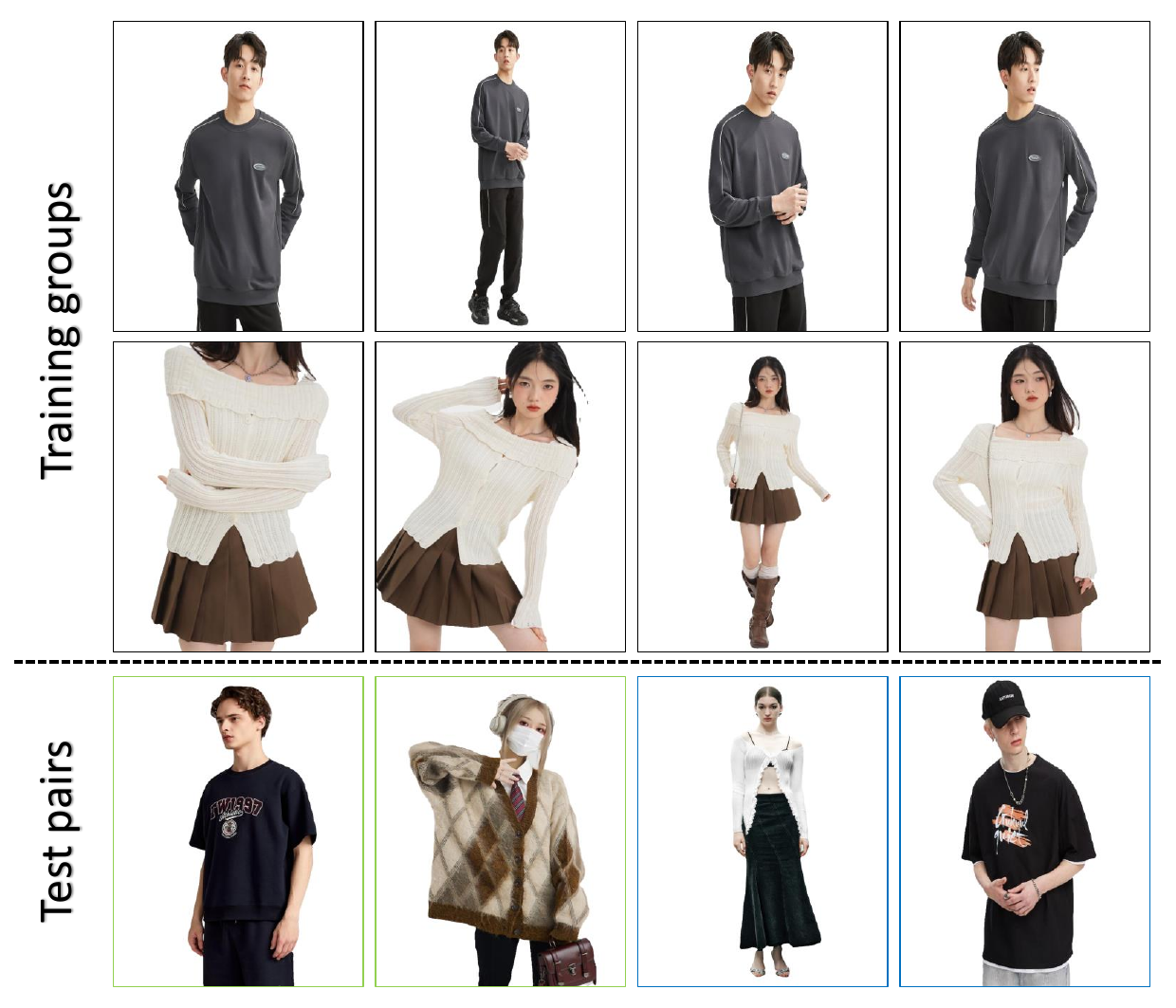}
  \caption{Some examples of our P2P-VTON dataset.}
  \label{fig:dataset_example}
\end{figure}
\subsection{P2P-VTON Dataset}
Our P2P-VTON dataset introduces a novel approach to VTON data collection, distinctly different from existing datasets such as VITON-HD and DressCode. Instead of the traditional pairing of flat garment images with single-worn instances, we provide comprehensive group images that capture the same person wearing identical garments across multiple poses. The dataset structure includes 2,787 groups with two images, 916 groups with three images, 316 groups with four images, 12 groups with twenty images, and one group with seven images. Each image in the dataset is accompanied by its corresponding flattened garment representation. For training purposes, we utilize intra-group image pairs, enabling the model to learn robust pose adaptation capabilities for garment transformation. The testing set comprises 1,786 pairs specifically curated to feature different individuals wearing distinct garments in various poses, providing a rigorous evaluation framework for assessing the model's generalization capabilities across diverse subjects and clothing items. This dataset structure offers unique advantages over existing alternatives. The availability of multiple views and poses for each garment-person combination facilitates effective training for person-to-person VTON tasks. Additionally, the significant pose variations within the dataset create challenging scenarios that push the boundaries of current VTON capabilities, as illustrated in Figure \ref{fig:dataset_example}.

\begin{figure*}[th]
  \includegraphics[width=\textwidth]{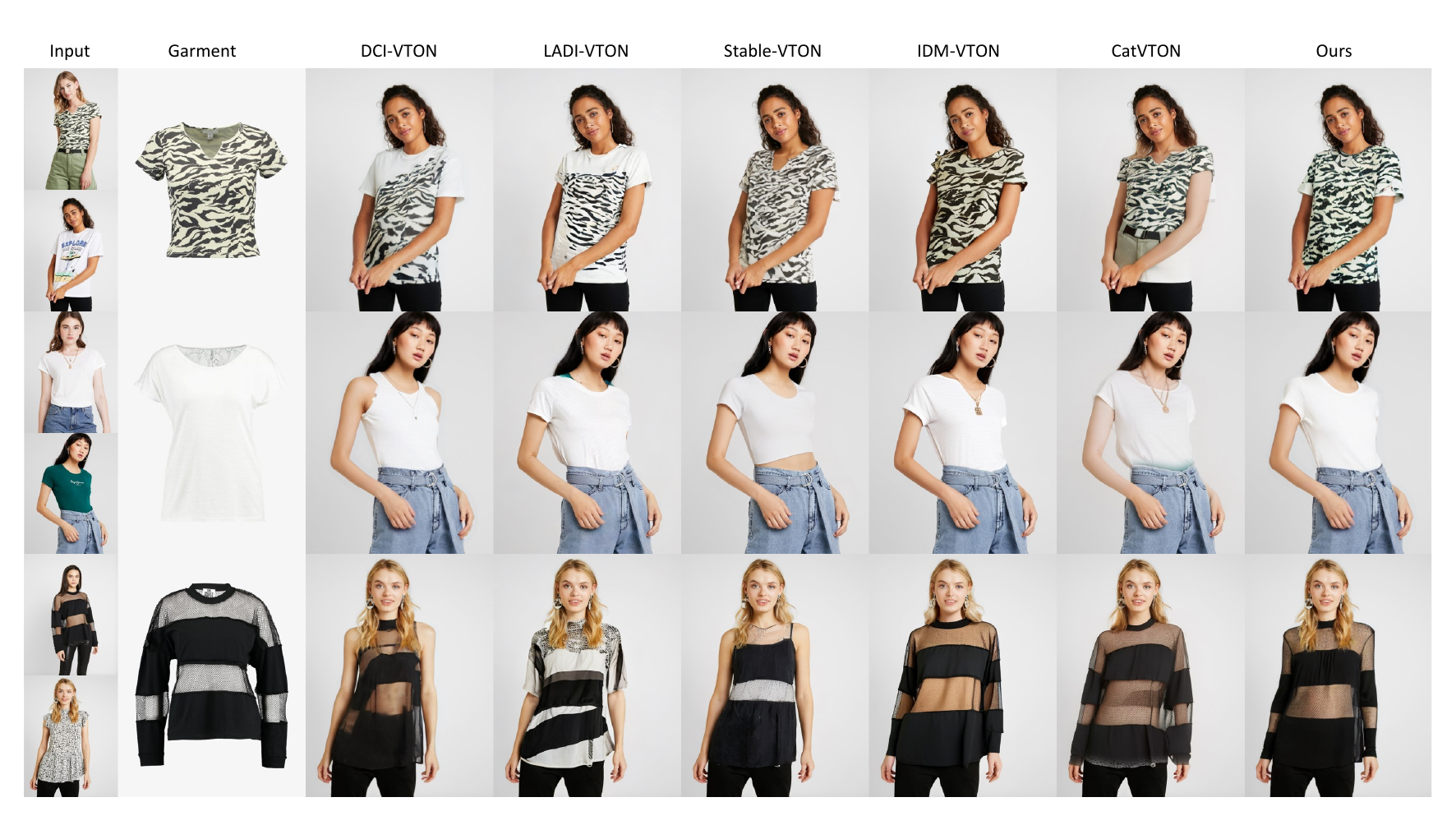}
  \caption{Qualitative results for person-to-person VTON task on the VITON-HD dataset.}
  \label{fig:tryon_baseline}
\end{figure*}
\begin{figure}[th]
  \includegraphics[width=1\columnwidth]{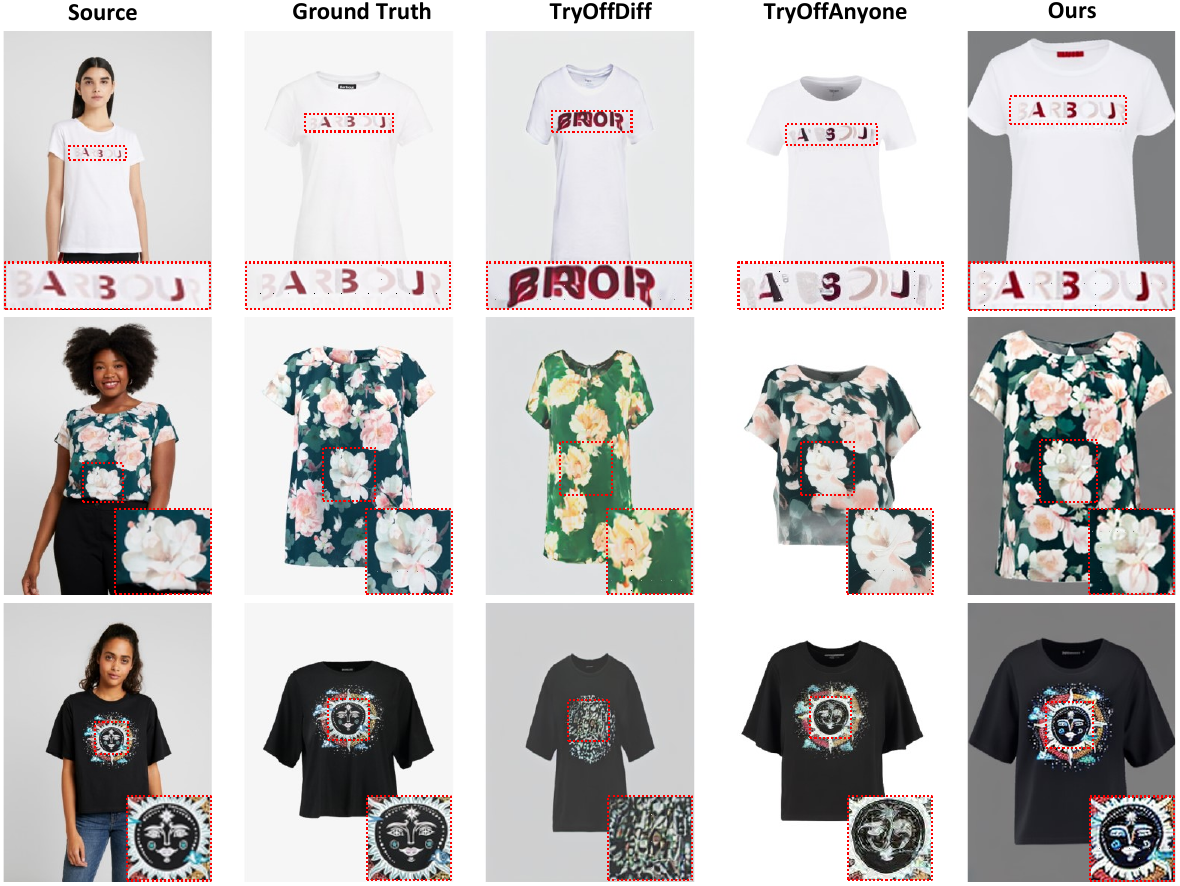}
  \caption{Qualitative results for the try-off subtask on the VITON-HD dataset. Due to the data augmentation applied during training, our results feature a gray background, with the garment filling the entire image.}
  \label{fig:tryoffbaseline}
\end{figure}
\section{Experiments}
\subsection{Experimental Setup}
\paragraph{Implementation details.}
All modules are trained using the Adam optimizer~\cite{loshchilov2017fixing} within the DDPM framework~\cite{ho2020denoising}, with a learning rate of 1e-5. Our training protocol includes specific data augmentation strategies for the flattening and warping modules, such as background neutralization (graying out) and garment scaling to maximize image coverage. Inference for all modules is conducted using DDIM~\cite{song2020denoising} with 20 sampling steps. For additional implementation details, please refer to the~\textit{Supplementary Materials}.

\paragraph{Datasets.}
We evaluate our method on the VITON-HD dataset~\cite{choi2021viton} and our newly proposed person-to-person (P2P) dataset. The VITON-HD dataset contains 13,679 image pairs, split into 11,647 training pairs and 2,032 test pairs. These images feature frontal-view photos of women wearing top garments, all at a resolution of $1024\times768$.
Our P2P VTON dataset comprises 15,144 training pairs and 1,786 test pairs of top garments, covering both men and women. It captures a diverse range of poses, body shapes, skin tones, and garments with varying textures and patterns. All images in the P2P VTON dataset maintain a resolution of $1024\times768$.

\paragraph{Compared Methods.}
On the VITON-HD dataset, we compare our FW-VTON with several state-of-the-art methods: DCI-VTON~\cite{gou2023taming}, LaDI-VTON~\cite{morelli2023ladi}, StableVITON~\cite{kim2024stableviton}, IDM-VTON~\cite{choi2025improving}, and CatVTON~\cite{chong2024catvton}. 
Except for CatVTON, we retrained or fine-tuned all other methods for consistency. As CatVTON’s training code is unavailable, we use the released checkpoint, which was trained on the VITON-HD dataset and includes the person-to-person task. This checkpoint is directly used for inference on the VITON-HD dataset. And CatVTON is excluded from the comparison on the P2P-VTON dataset.
In addition, we compared the flattening module of our proposed method against TryOffDiff~\cite{velioglu2024tryoffdiff} and TryOffAnyone~\cite{xarchakos2024tryoffanyone} on the VITON-HD dataset. Although FW-VTON supports training and testing at a resolution of $1024\times768$, all experiments were conducted at $512\times384$ resolution to ensure a fair comparison.

\paragraph{Evaluation Metrics.}
In the person-to-person VTON task, as our test scenarios involve different garments worn by different individuals, ground truth is unavailable for the generated results. Therefore, we employ Fréchet Inception Distance (FID)~\cite{binkowski2018demystifying} and Kernel Inception Distance (KID)~\cite{parmar2022aliased} to measure the distribution similarity between synthesized and real samples. Additionally, we created composite images using various methods for 200 randomly selected pairs from our test set, each at a resolution of 512×384. These images were evaluated by a panel of 50 human judges, who were tasked with identifying the method they believed most effectively performed the task. For the flattening module comparisons, we use multiple metrics: Structural Similarity Index (SSIM)~\cite{wang2004image}, Learned Perceptual Image Patch Similarity (LPIPS)~\cite{zhang2018unreasonable}, DISTS~\cite{ding2020image}, FID, and KID.

\begin{table}[!t]%
\caption{Quantitative results of the person-to-person VTON task on the VITON-HD dataset. The user study scores represent the selection proportions, with the total sum equal to 100. KID values have been multiplied by 1000.}
\label{tab:one}
\begin{minipage}{\columnwidth}
\begin{center}
\begin{tabular}{lrrrr}
  \toprule
    Method           & FID$\downarrow$   & KID$\downarrow$ & User Study$\uparrow$ \\ \midrule
    DCI-VTON        & 15.31 & 6.81 & 0.42 \\
    LaDI-VTON        & 12.77  & 4.33 & 0.21  \\
    Stable-VTON       & 8.86  & 1.03  & 9.73  \\
    IDM-VTON        & 9.12   & 0.99 &  10.12 \\
    CatVTON        & 10.25  &  1.87 & 13.28 \\
    \textbf{Ours(FW-VTON)}           & \textbf{8.53} & \textbf{0.83} & \textbf{66.24}\\ 
  \bottomrule
\end{tabular}
\end{center}
\bigskip\centering

\end{minipage}
\end{table}%

\begin{table}[!t]%
\caption{Quantitative results of the person-to-person VTON task on our proposed P2P-VTON dataset.}
\label{tab:two}
\begin{minipage}{\columnwidth}
\begin{center}
    \begin{tabular}{lrrrrr}
    \toprule
    Method           & FID$\downarrow$   & KID$\downarrow$ & User Study$\uparrow$  \\ \midrule
    DCI-VTON        & 26.49 & 7.11 & 1.96\\
    LaDI-VTON        & 21.50  & 3.88 & 2.52\\
    Stable-VTON       & 19.01  & 1.81 & 10.77\\
    IDM-VTON        & 19.35   & 1.87 & 23.05\\
    \textbf{Ours(FW-VTON)}           & \textbf{18.78} & \textbf{1.55} & \textbf{61.70}\\ 
    \bottomrule
\end{tabular}
\end{center}
\bigskip\centering

\end{minipage}
\end{table}%
\subsection{Experimental Results}
\begin{figure*}[t]
\centering 
\includegraphics[width=0.8\linewidth]{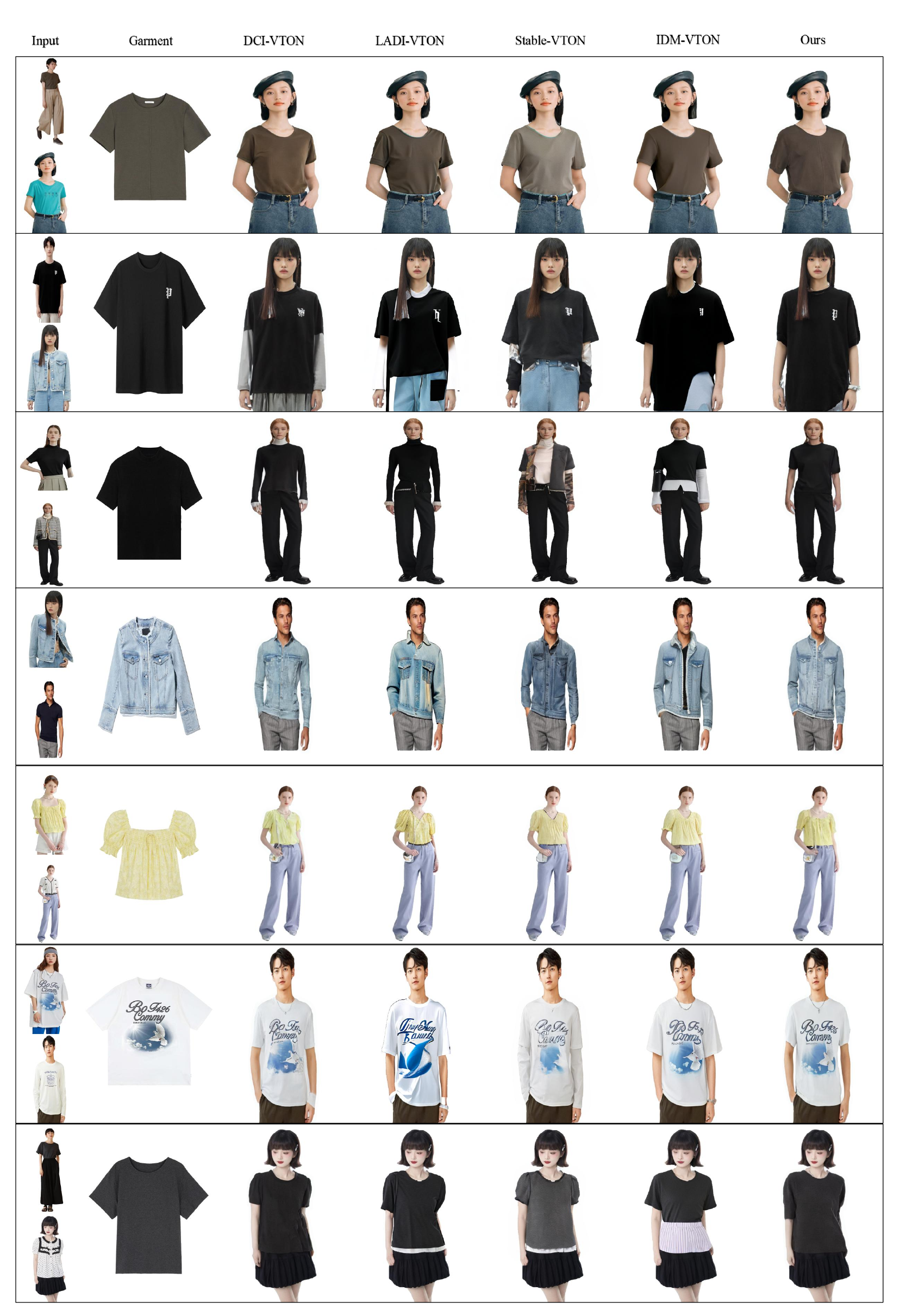} 
\caption{Qualitative results for the person-to-person VTON task on our P2P-VTON dataset. Please note that the garment image is not available in our task and is included here solely for comparison purposes.} 
\label{fig:tryon_baseline_2} 
\end{figure*}

\paragraph{Qualitative Results.}
Figure~\ref{fig:tryoffbaseline} presents examples from our flattening module, TryOffDiff, and TryOffAnyone on the VITON-HD test set. TryOffDiff, which relies solely on an image adaptor, fails to produce satisfactory results. TryOffAnyone, which directly concatenates all condition information into the U-Net input channels, performs reasonably well but struggles with text rendering, fine pattern preservation, and completing occluded garment regions. In contrast, our flattening module, based on a dual U-Net architecture with explicit pose conditioning, achieves significantly more accurate and detailed results.
Figure~\ref{fig:tryon_baseline} shows comparisons between FW-VTON and competing methods on the VITON-HD dataset (P2P-VTON results are provided in Figure~\ref{fig:tryon_baseline_2}). While methods such as Stable-VTON, IDM-VTON, and CatVTON outperform earlier approaches like DCI-VTON and LaDI-VTON in preserving garment details and maintaining image consistency, they still struggle with fine-grained visual quality—an area where our method excels.
Notably, CatVTON, which processes two images directly through concatenation, occasionally blends unrelated content, such as altering the target person’s skin tone to match the source, or misidentifying garment regions (\textit{e.g.}, transferring pants or accessories like necklaces). Similar artifacts are observed in IDM-VTON. Our method avoids these issues by removing accessories during the flattening stage and generating a clean, garment-focused representation, leading to better garment alignment and more realistic results.

\begin{table}[t]%
\caption{Quantitative results for the try-off subtask on the VITON-HD dataset.}
\label{tab:three}
\begin{minipage}{\columnwidth}
\begin{center}
    \begin{tabular}{lrrrrr}  
    \toprule
    Method           & SSIM$\uparrow$  & LPIPS$\downarrow$ & FID$\downarrow$   & KID$\downarrow$  & DISTS$\downarrow$  \\ \midrule
    TryOffDiff        & 0.617 & 0.486 & 21.156 & 7.3 &   0.2522\\
    TryOffAnyone        & 0.716  & 0.441  & 15.320  & 4.2  & 0.2433  \\
    \textbf{Ours}             & \textbf{0.739} & \textbf{0.363} & \textbf{10.043}  & \textbf{1.7} &    \textbf{0.2016}\\ 
    \bottomrule
\end{tabular}
\end{center}
\bigskip\centering
\end{minipage}
\end{table}%

\paragraph{Quantitative Results.}
Table~\ref{tab:three} reports results for the try-off subtask on the VITON-HD dataset, where our flattening module surpasses all baselines. Table~\ref{tab:one} shows person-to-person VTON performance on VITON-HD. While CatVTON achieves strong results on garment-to-person VTON using a single U-Net with concatenated inputs~\cite{chong2024catvton}, IDM-VTON and Stable-VTON obtain better results on the more challenging person-to-person task by introducing reference encoders or Stable Diffusion modules.
Our approach further improves upon IDM-VTON by decomposing the VTON task into three subtasks, enabling more effective use of condition information. As shown in Table~\ref{tab:two}, performance on our proposed P2P-VTON dataset—which includes greater garment variety and more complex poses—is lower overall compared to VITON-HD, yet our method consistently outperforms all baselines across evaluation metrics, demonstrating its robustness in more challenging scenarios.

\begin{figure}[t]
  \includegraphics[width=\columnwidth]{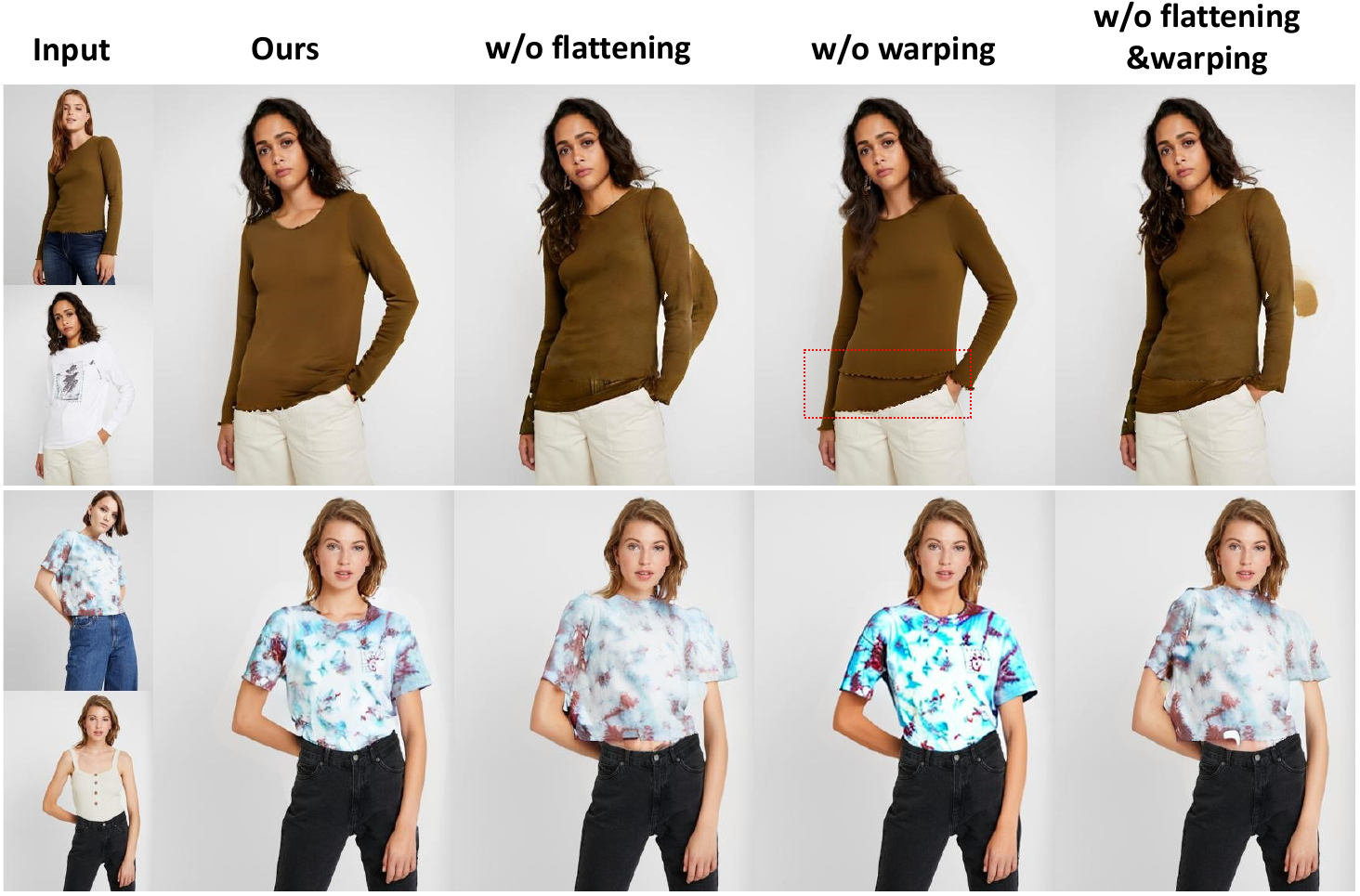}
  \caption{Ablation examples of our two modules. Note that in the person-to-person VTON task, the garment image is unavailable.}
  \label{fig:ablation_two_modules}
\end{figure}

\begin{table}[t]%
\caption{Ablation quantitative results on the VITON-HD dataset.}
\label{tab:four}
\begin{minipage}{\columnwidth}
\begin{center}
    \begin{tabular}{lrrrrr}
    \toprule
    Method & FID$\downarrow$   & KID$\downarrow$  \\ 
    \midrule
    (a) w/o Flattening w/o Warping       & 12.68 & 3.81 \\
    (b) w/o Warping       & 9.57 &  1.32 \\
    (c) w/o Flattening  &  11.90 & 3.10 \\
    (d) Ours  & \textbf{8.53} & \textbf{0.83} \\
    \bottomrule
\end{tabular}
\end{center}
\bigskip\centering
\end{minipage}
\end{table}%
\subsection{Ablation Studies}
Here, we conduct ablation studies to demonstrate the effectiveness of our flattening and warping modules. For additional ablation studies, please refer to the~\textit{Supplementary Materials}.
\paragraph{Effect of Flattening and Warping Modules.}
We conduct separate evaluations to assess the contributions of the flattening and warping modules in our framework.
For the flattening module, we replace the flattened garment representation $\mathcal{G}_f$ with the raw conditional garment image $\mathcal{G}_c$ during both training and inference. For the warping module, we substitute our warping network with that of PFAFN~\cite{ge2021parser}, a widely used method in prior VTON works.
As illustrated in Figure~\ref{fig:ablation_two_modules}, the absence of our flattening module leads to incomplete garment representation. This results not only in a loss of garment details but also in noticeable structural inconsistencies. Similarly, replacing our warping module with PFAFN yields inaccurate warped guidance $\mathcal{G}_w$, often causing structural errors such as hallucinated or misplaced garment parts.
Quantitative results in Table~\ref{tab:four} further confirm the importance of both modules, demonstrating clear performance degradation when either component is ablated.

\section{Conclusion}
Compared to previous VTON methods, which primarily focus on the garment-to-person VTON task, this paper addresses the more challenging person-to-person VTON scenario. While recent VTON methods employ dual U-Net or simplified U-Net architectures to tackle the task, we argue that for complex scenarios like person-to-person VTON, decomposing the process into three stages—flattening, warping, and integration—leads to more robust and higher-quality results.
Furthermore, we note that recent works focusing on the try-off task align closely with the functionality of our flattening module. By leveraging a dual U-Net architecture and incorporating well-designed conditional information, our method achieves significant improvements over the state-of-the-art performance.

\bibliographystyle{ACM-Reference-Format}
\bibliography{sample-bibliography}

\newpage

\begin{figure*}[t]
\centering 
\includegraphics[width=0.8\linewidth]{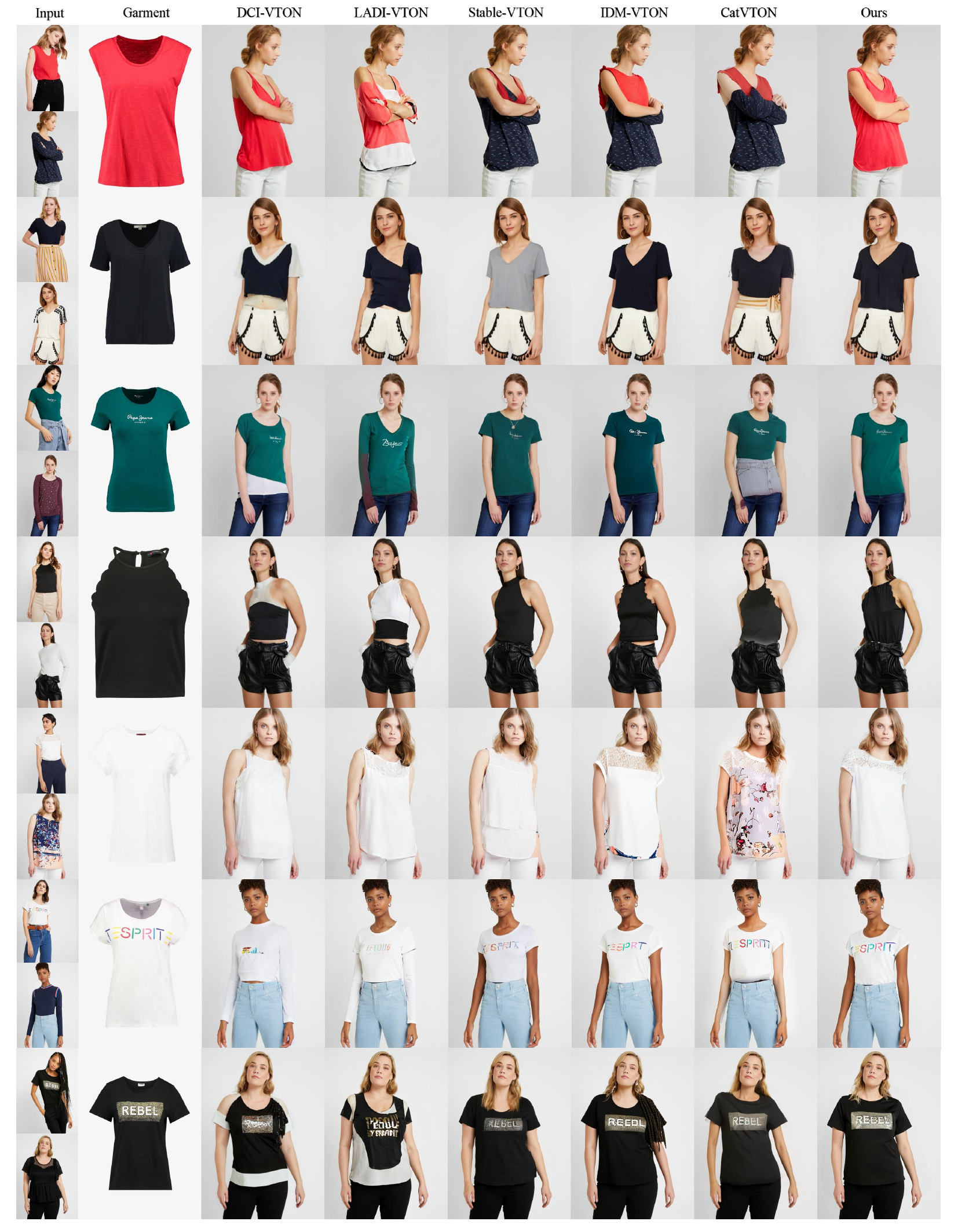} 
\caption{More qualitative results for the person-to-person VTON task on the VITON-HD dataset.} 
\label{fig:Appendix_VITON} 
\end{figure*}

\clearpage
\appendix










\section{Supplementary Materials of FW-VTON: Flattening-and-Warping for Person-to-Person Virtual Try-on}
This section provides additional details about the implementation of our methods and the compared approaches.

\paragraph{DCI-VTON}
In our person-to-person VTON setting, flat garment images are unavailable. Therefore, we replace them with cut garment images extracted from the source image in the DCI-VTON~\cite{gou2023taming} pipeline. For the VITON-HD dataset, which does not include garment images in different poses, we bypass training the warping module. Instead, we use the pre-trained warping module for inference and only train the final diffusion model. Conversely, for the P2P-VTON dataset, which includes garment images in various poses, we train the warping module first, followed by training the final diffusion model. 

\paragraph{LaDI-VTON}
For the VITON-HD dataset, we handle LaDI-VTON \cite{morelli2023ladi} in the same way as DCI-VTON. Specifically, we use the pre-trained warping module for inference and only fine-tune the final diffusion model. For the P2P-VTON dataset, we first train the warping module, followed by fine-tuning the Enhanced Mask-Aware Skip Connection (EMASC) modules for 40 epochs. We then train the inversion adapter to obtain its checkpoint, which, along with the released diffusion checkpoint, is used to fine-tune the diffusion model on the P2P-VTON dataset for 40 epochs.

\paragraph{Stable-VITON}
In the Stable-VITON~\cite{kim2024stableviton} pipeline, the Stable Diffusion~\cite{rombach2022high} encoder serves as the garment condition extractor, integrated into the denoising U-Net via a zero cross-attention block. This block significantly increases training complexity; training from scratch requires 1,000 epochs, taking nearly a week on an 8×A800 GPU setup. So for the VITON-HD dataset, we fine-tune the model from the released checkpoint, pre-trained on VITON-HD. The fine-tuning process, which adapts the model from flat garment images to cut garment images, involved 40 epochs. Similarly, in FW-VITON training, the final integration diffusion module was also trained for 40 epochs, ensuring a fair comparison. For the P2P-VTON dataset, we fine-tune the model for 40 epochs starting from the released checkpoint, using a learning rate of 1e-5. 

\paragraph{IDM-VTON}
For the VITON-HD dataset, we retrain IDM-VTON~\cite{choi2025improving} for 130 epochs as specified in the original paper, replacing flat garment images with cut garment images. For the P2P-VTON dataset, we train the model for 130 epochs as well. However, due to the lack of detailed garment captions in our P2P-VTON dataset, we used a generic label, ``Upper Clothes'', as the garment caption. 

\paragraph{CatVTON}
Since CatVTON’s~\cite{chong2024catvton} training code is unavailable, we utilized the released checkpoint, which was trained on the VITON-HD dataset, ensuring that the input configuration matched the original training setup. We performed inference directly on the VITON-HD dataset using these checkpoint. CatVTON is excluded from comparisons on the P2P-VTON dataset. 

\paragraph{FW-VTON}
In our proposed FW-VTON framework, the warping, flattening, and integration modules are initialized with weights from the publicly released SD-Image-Variations-Diffusers checkpoint. Each module (warping, flattening, and integration) is trained separately for 45,000 steps using 8×NVIDIA A800 GPUs, with each training process taking approximately 6 hours.
A key difference in training between the VITON-HD and P2P-VTON datasets lies in the construction of the warping module. Since the VITON-HD dataset does not provide garment images under diverse poses, we adopt the PFAFN warping module to synthesize warped garment images from flat garment inputs, conditioned on the target person's pose. These synthesized pairs are then used as training data for our warping module.
During inference, all modules use 20 sampling steps with DDIM~\cite{song2020denoising}, achieving an average processing time of approximately 3 seconds per image on a single A800 GPU.

\begin{table*}[t]
\centering
\setlength{\tabcolsep}{3pt}
\begin{tabular}{crrrrr}
\toprule
Method          & SSIM$\uparrow$  & LPIPS$\downarrow$ & FID$\downarrow$   & KID$\downarrow$  & DISTS$\downarrow$  \\ 
\midrule
Add $\mathcal{P}_m$ with noise       & 0.8137 & 0.2404 & 13.17 & 3.8 &   0.1697\\
Concatenate $\mathcal{P}_m$ with noise   & 0.7517 & 0.3837 & 17.30 & 5.8 &   0.2191 \\
\textbf{Concatenate $\mathcal{P}_m$ with $\mathcal{G}_c$}  & \textbf{0.8261} & \textbf{0.2180} & \textbf{10.97}  & \textbf{2.4} &    \textbf{0.1684}\\
\bottomrule
\end{tabular}
\caption{Ablation study of concatenating $\mathcal{P}_m$ with $\mathcal{G}_c$ in Warping Module. $\mathcal{G}_c$ denotes the cut garment image from source person image, $\mathcal{P}_m$ denotes the corresponding garment mask region in target person image.}

\label{tab:Concatenate}
\end{table*}

\begin{figure}[t] \centering \includegraphics[width=1\linewidth]{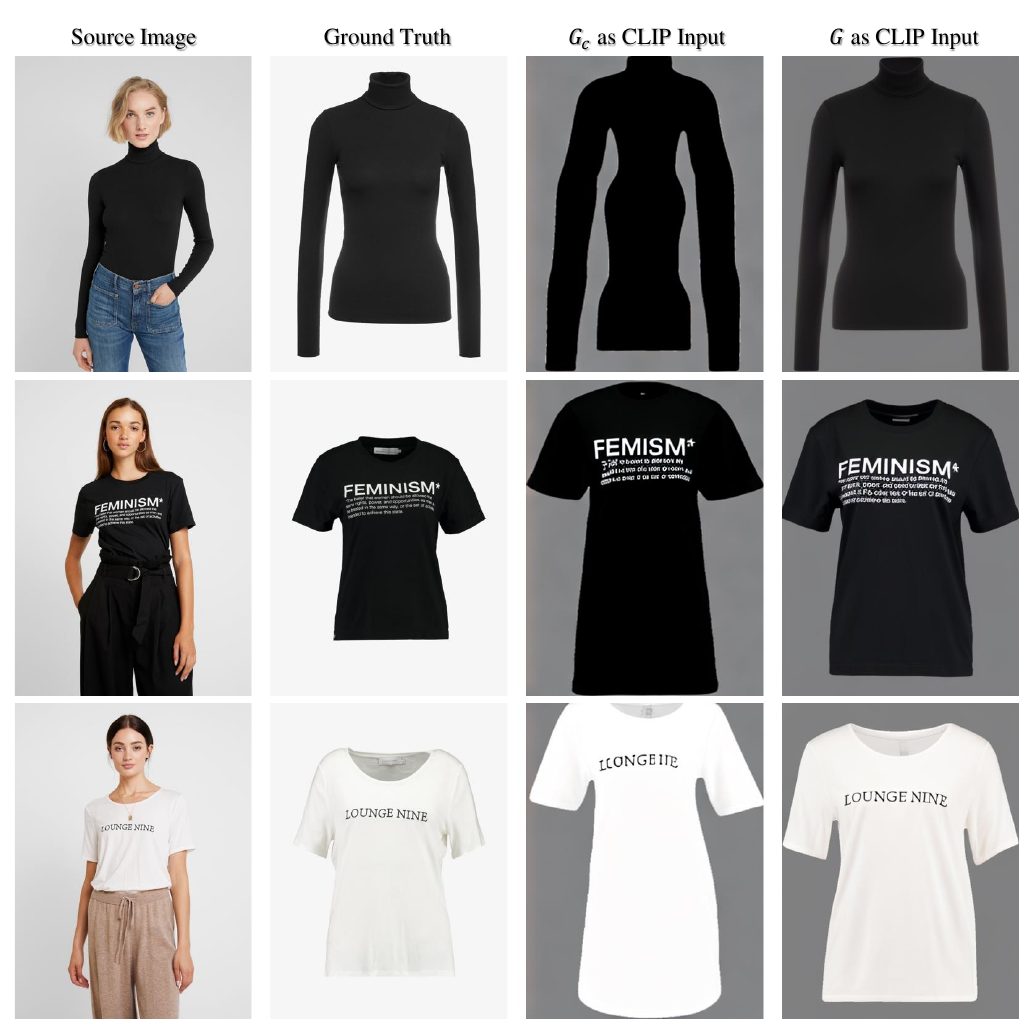} \caption{Ablation examples of $\mathcal{G}$ as CLIP input.} 
\label{fig:ablation_condition} \end{figure}

\begin{table}[t]
\centering
\setlength{\tabcolsep}{3pt}
\begin{tabular}{crrrrr}
\toprule
Method          & SSIM$\uparrow$  & LPIPS$\downarrow$ & FID$\downarrow$   & KID$\downarrow$  & DISTS$\downarrow$  \\ 
\midrule
$\mathcal{G}_c$ as CLIP input       & 0.658 & 0.399 & 19.71 & 6.6 &   0.2251\\
\textbf{$\mathcal{G}$ as CLIP input}    & \textbf{0.739} & \textbf{0.363} & \textbf{10.04}  & \textbf{1.7} &    \textbf{0.2016}\\
\bottomrule
\end{tabular}
\caption{Ablation study of CLIP input in the flattening module. $\mathcal{G}$ denotes the original garment image, and $\mathcal{G}_c$ represents the cut garment region from $\mathcal{G}$.}

\label{tab:G importance}
\end{table}

\section{More Ablation Studies}
\subsection{Effect of $\mathcal{G}$ in Flattening Module}
While the reference net effectively extracts garment information, the key challenge lies in enabling the denoising U-Net to better reconstruct the original garment image. Our experiments reveal that simply providing the garment-wearing image $\mathcal{G}$ to the CLIP encoder greatly enhances the results (see Table~\ref{tab:G importance}). This addition improves the model's consistency with the ground truth and enhances the generation of finer details, as shown in Figure~\ref{fig:ablation_condition}.

\subsection{Effect of Concatenating $\mathcal{P}_m$ with $\mathcal{G}_c$ in Warping Module}
In our warping module, effectively warping the garment to align with the target pose requires determining the optimal method for integrating the cut garment image, $\mathcal{G}_c$, with the corresponding garment mask region from the target person image, $\mathcal{P}_m$. We explored three different approaches:
\begin{enumerate}
    \item Directly adding the mask with noise to the input channel of the denoising U-Net: We encode $\mathcal{P}_m$ using a Variational Autoencoder (VAE)~\cite{kingma2013auto}, and then modify the initial noise by adding the mask value. This approach effectively translates the task as an image variation problem, where the image evolves from the mask to the warped garment.
    
    \item Concatenating the mask with the noise in the input channel of the denoising U-Net: In this method, we resize $\mathcal{P}_m$ and concatenate the mask with the noise, increasing the input channel size of the denoising U-Net from 4 to 5 by adding zero parameters.
    
    \item Concatenating the mask with $\mathcal{G}_c$ in the input channel of the reference net: Similarly, we resize $\mathcal{P}_m$ and concatenate the mask with $\mathcal{G}_c$, modifying the input channel size of the reference network from 4 to 5 by adding zero parameters.
\end{enumerate}
We compare the performance on the VITON-HD dataset, where the warping ground truth for both training and testing is generated using the warping module from PFAFN~\cite{ge2021parser}. The warping result metrics are summarized in Table~\ref{tab:Concatenate}. As shown, the third setting—concatenating $\mathcal{P}_m$ with $\mathcal{G}_c$ in the reference net’s input channel—achieves the best performance. This method effectively incorporates garment mask information to enhance the warping process, outperforming the other two settings.

\begin{figure*}[t]
\centering 
\includegraphics[width=0.8\linewidth]{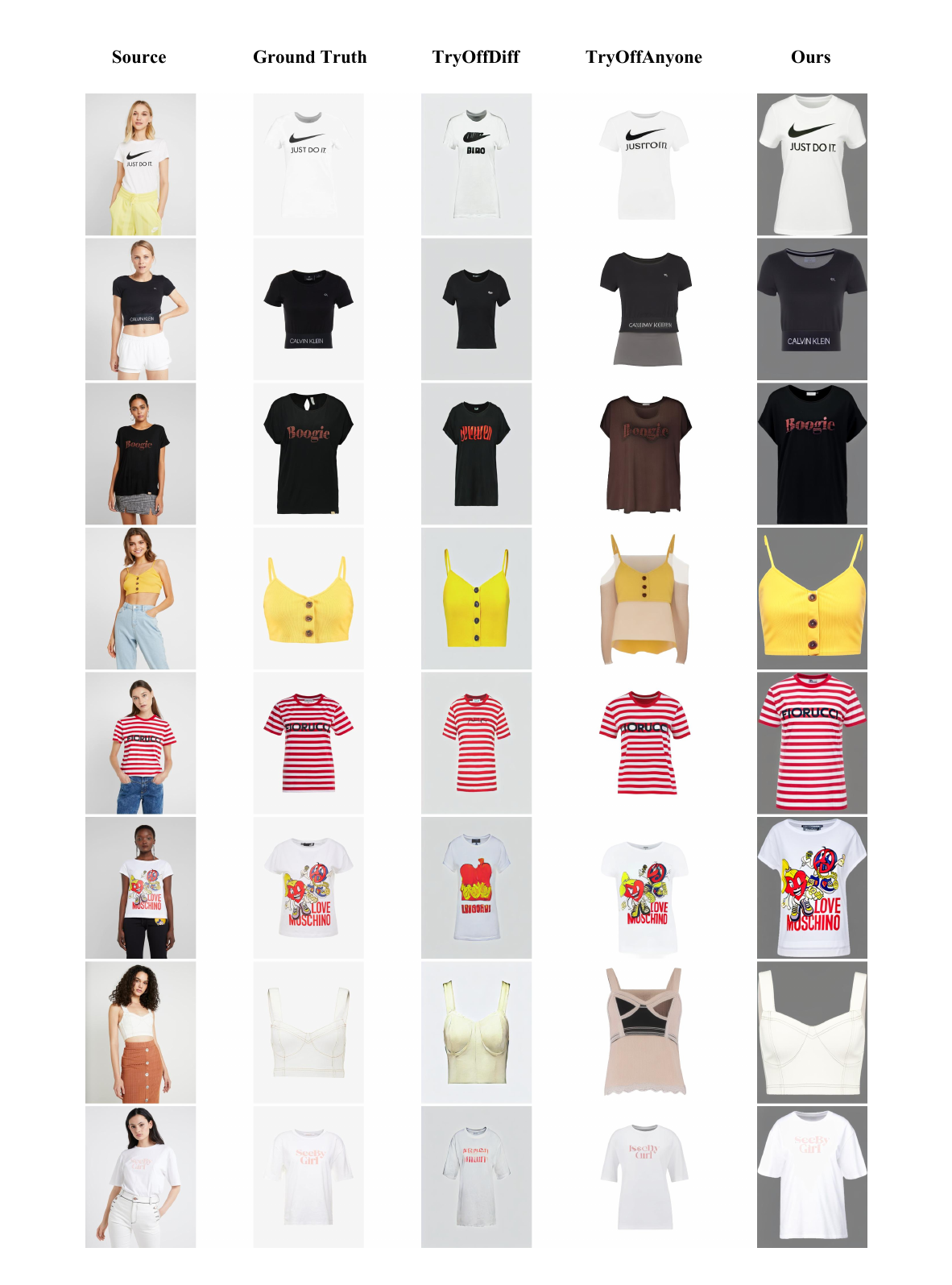} 
\caption{More qualitative results for the try-off subtask on the VITON-HD dataset. Due to the data augmentation applied during training, our results feature a gray background, with the garment filling the entire image.} 
\label{fig:Appendix_vtoff} 
\end{figure*}

\begin{table}[t]
\centering
\setlength{\tabcolsep}{3pt}
\begin{tabular}{crrr}
\toprule
Try-off Method     & Try-on Method & FID & KID   \\ 
\midrule
\multicolumn{1}{c}{} & DCI-VTON     & 16.62 & 9.34 \\
TryoffDiff           & LaDI-VTON    & 14.84 & 8.41 \\
                     & Stable-VTON  & 11.15 & 3.98 \\
                     & IDM-VTON     & 11.72 & 2.53 \\
\midrule
\multicolumn{1}{c}{} & DCI-VTON     & 14.18 & 7.01 \\
TryoffAnyone           & LaDI-VTON    & 14.77 & 8.67 \\
                     & Stable-VTON  & 11.12 & 3.92 \\
                     & IDM-VTON     & 10.67 & 2.25 \\
\bottomrule
\end{tabular}
\caption{Ablation study of CLIP input in the flattening module. $\mathcal{G}$ denotes the original garment image, and $\mathcal{G}_c$ represents the cut garment region from $\mathcal{G}$.}

\label{tab:Tryoff+Tryon}
\end{table}

\begin{table}[t]
\centering
\setlength{\tabcolsep}{3pt}
\begin{tabular}{crr}
\toprule
Method        & FID$\downarrow$   & KID$\downarrow$  \\ 
\midrule
w/ Cross-Attention       & 8.74 & 1.06  \\
\textbf{w/o Cross-Attention(FW-VTON)}    & \textbf{8.53} &    \textbf{0.83}\\
\bottomrule
\end{tabular}
\caption{Ablation study of whether skip the cross-attention layer in dual U-Net.}

\label{tab:skip_cross}
\end{table}

\subsection{Effect of Skipping the Cross-Attention Layer in the Integration Module}

The cross-attention layer in Stable Diffusion~\cite{rombach2022high} is primarily designed to align image and text modalities. 
In the context of virtual try-on, hence, the cross-attention is expected to contribute structural guidance rather than semantic alignment. 
Since our integration module explicitly injects the warped garment image $\mathcal{G}_w$ into the denoising U-Net input, we hypothesize that the cross-attention layer becomes redundant. 
To validate this hypothesis, we compare two variants of our integration module:
\begin{itemize}
    \item w/o Cross-Attention: Our default setting, which removes all cross-attention layers in both the reference and denoising U-Nets (as described in the Method section of the main paper).
    \item w/ Cross-Attention: A variant where we retain the cross-attention layers and use a CLIP encoder~\cite{radford2021learning} to encode the flattened garment image $\mathcal{G}_f$. The resulting CLIP embedding is then used as the context input for cross-attention in both the reference and denoising U-Nets.
\end{itemize}
Both versions are trained and evaluated using the same $\mathcal{G}_f$ and $\mathcal{G}_w$ inputs for fair comparison. As shown in Table~\ref{tab:skip_cross}, removing the cross-attention layers improves the final result and, due to the simplified architecture, also accelerates inference.


\appendix

\end{document}